\documentclass[letterpaper,twocolumn,10pt]{article}
\usepackage{usenix,epsfig,endnotes}
\usepackage{graphicx, multirow}
\usepackage[labelfont=bf]{caption}
\usepackage[table]{colortbl}
\usepackage[mathscr]{euscript}
\DeclareSymbolFont{rsfs}{U}{rsfs}{m}{n}
\DeclareSymbolFontAlphabet{\mathscrsfs}{rsfs}
\begin{document}

\date{December 2022}

%make title bold and 14 pt font (Latex default is non-bold, 16 pt)
\title{\Large \bf Graph Neural Operators for Classification of Spatial Transcriptomics Data}

\author{
{\rm Junaid Ahmed}\\
Johns Hopkins University \\
BlueLightAI Inc. \\
E-mail: \href{mailto:junaid.ahmed@bluelightai.com}{junaid.ahmed@bluelightai.com}
% }
\and
{\rm Alhassan S. Yasin}\\
Johns Hopkins University\\
\href{https://yasinresearch.com/}{Yasin Research Group}\\
E-mail: \href{mailto:ayasin1@jhu.edu}{ayasin1@jhu.edu}
}

\maketitle

\subsection*{Abstract}
The inception of spatial transcriptomics has allowed improved comprehension of tissue architectures and the disentanglement of complex underlying biological, physiological, and pathological processes through their positional contexts. Recently, these contexts, and by extension the field, have seen much promise and elucidation with the application of graph learning approaches. In particular, neural operators have risen in regards to learning the mapping between infinite-dimensional function spaces. With basic to deep neural network architectures being data-driven, i.e. dependent on quality data for prediction, neural operators provide robustness by offering generalization among different resolutions despite low quality data. Graph neural operators are a variant that utilize graph networks to learn this mapping between function spaces. The aim of this research is to identify robust machine learning architectures that integrate spatial information to predict tissue types. Under this notion, we propose a study incorporating various graph neural network approaches to validate the efficacy of applying neural operators towards prediction of brain regions in mouse brain tissue samples as a proof of concept towards our purpose. We were able to achieve an F1 score of nearly 72\% for the graph neural operator approach which outperformed all baseline and other graph network approaches.

\section{Introduction}

A neural network takes in a single input and returns a single output. However, this is not sufficient in capturing systematic relationships by way of the spatial domain. A neural operator~\cite{NeuralOp} takes in a function and outputs a function by way of partial differential equations. These are effective in modeling complex systematic relationships and may be specifically applicable towards interpolation and segmentation tasks. Neural operators seem promising towards spatial transcriptomics not just in determining the items present in the cell, but actual spatial positions of those items as well.

To obtain a fundamental understanding of neural operators, we consider that the spatial transcriptomics data can be viewed as a continuous function as opposed to a static set of at-time collections. Furthermore, we view the samples of data as being discretized from the continuous function that is defined on the spatial domain $\mathcal{D}$ of each tissue slide. Neural operators are essentially built on the premise of partial differential equations, and therefore are designed to work with discretized functions.

The key part of a neural operator is defining a kernel layer that works on a function with learnable parameters. Controlling the properties of the kernel function ultimately controls the layer. There is quite a bit of research to determine optimal kernel functions, for example through the implementation of neural networks and transformers. Additionally, there is some effort being dedicated towards determining a universal approximation kernel function. For the aim of this research, we have identified two kernel functions: a simple kernel involving the Euclidean spatial coordinates, and a graph network based kernel layer. 

The equation below describes the simple kernel for the spatial neural operator.

\[ u(x) = \sigma\Big(W_{v_{t}}(x) \sum_{y \in N}\kappa_{\phi}(x,y) v_{t}(y)\Big) \]

For this operator, $\kappa(x,y)$ represents the neural network which accepts the Euclidean coordinates as inputs and outputs an NxN matrix. $W$ is a linear transformation that acts on $v$. $\sigma$ was a constant value for manipulating the kernel that was not parameterized.

The equation below illustrates the general idea of a graph neural operator, or GraphPDE~\cite{GraphPDE}.

\[ v_{t+1}(x) = \sigma\Big(W_{v_{t}}(x) + \frac{1}{|N(x)|}\sum_{y \in N}\kappa_{\phi}(\epsilon(x,y)) v_{t}(y)\Big) \]

For graphs constructed on the spatial domain $\mathcal{D}$, the latter portion of this equation is the same as the message passing aggregation of graph neural networks in accordance with the edge attributes. So simply, the graph kernel is a an aggregation of messages. Again here $W$ is a linear transformation that acts on $v$ and $\sigma$ is an activation function. The additional caveats we consider are $v_t$ as the node features, $e(x,y)$ as the edge features, and $N(x)$ as the neighborhood of x according to the graph. $\kappa_{\phi}(e(x,y))$ is indicative of the neural network which takes graph edge features as inputs (as opposed to Euclidean coordinates of the spatial neural operator) and outputs an NxN matrix.

\section{Materials and Methods}
To validate that our materials were prudent for analysis, we established the baseline models we intended to use for evaluating our filtered data. The baseline models included a Logistic Regression Classifier, a neural network, a random forest, and an XGBoost classifier [Table~\ref{tab:Abbs}]. These were not considered experimental models because, for one, they have been pre-established in industry as best suited techniques in the scope of classification tasks, and second, our aim is to validate graph network approaches as viable state of the art classification methods. %The process required much trial and error until we were able to establish the final analysis set. The steps of this iterative process are described below.

\subsection{Dataset}

The goal is to select an optimal spatial transcriptomics dataset. We plan to pursue a supervised classification task to validate the efficacy of using state of the art graph network approaches towards tissue type prediction. For a dataset to be considered towards our intended research, one requirement is for the dataset to have plentiful collected spots across a large feature set of genes. Second, it must have been gathered across a plethora of tissue samples through robust profiling techniques. It is generally difficult to find datasets in this domain which satisfy all of these requirements. Datasets may not be well suited for supervised classification by either the lack of a distinctive target feature(s), or the more prevalent case is that there aren't enough tissue samples to decipher a decent generalization.
% One key requirement for determining  viable spatial transcriptomics sets for our research is that the data points should be representative of  a large amount of tissue samples.

\begin{figure}[htp]
    \centering
    \includegraphics[width=9cm]{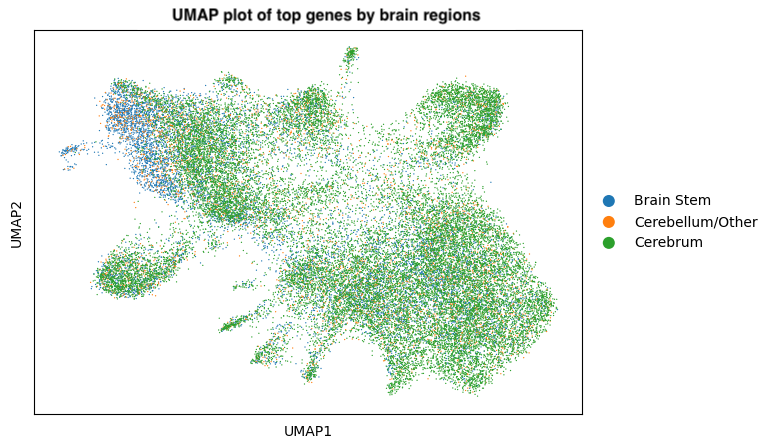}
    \caption{UMAP visualization constructed from the subsampled and normalized gene set ($n\approx232$) across all spots where colors are indicative of the associated brain region. The plot shows strong imbalance in the dataset where a significant number of spots are localized in the Cerebrem as opposed to the other regions. This will be strongly considered towards balanced data preparation for modeling.}
    \label{fig:umap}
\end{figure}

We used a central repository of curated spatial transcriptomics datasets~\cite{STOmicsDB} in order to find suitable datasets as per our requirements. The mouse brain atlas dataset~\cite{Pubmed} we selected for our research contains about 34,000 spots and 23,000 genes with 75 samples. This satisfies our requirement of usable datasets by way of number of tissue sections and having been profiled by Illumina spatial transcriptomics profiling techniques. Captured within are the gene expression signatures which define the spatial organization of molecularly discrete subregions. Figure~\ref{fig:umap} illustrates how the dataset is a “molecular atlas”~\cite{Pubmed}, curated to firstly define the identity of brain regions, as well as establish a molecular code for the mapping and targeting of discrete neuroanatomical domains.

%We intend to utilize this dataset towards a supervised learning approach for prediction of brain regions given new spatial information. 

Normalization is a key component to preparing this data before analysis. We required a strategy to distill the large feature set to a trainable set. Additionally, we discovered that introducing sparsity, as per their raw counts set, was not helpful in differentiating variable genes. Fortunately, the authors had provided a pre-normalized set that facilitated our discoveries.

\subsection{Exploratory Data Analysis}

%Our initial approach involved using the raw counts and metadata sets to begin our analyses. In order to distill our feature set to a workable group, we arbitrarily filtered it to a set of the top 200 most variable genes. We soon understood through our exploratory data analysis that this was a naive approach and that a convoluted approach would be necessary in order to obtain the most relevant and variable genes.

%We hearkened back to the original paper for the dataset to gain inspiration. 

Following the supplemental information provided by the authors of our intended dataset, we discovered that the authors had already gone through a sophisticated filtration step~\cite{Pubmed} which allowed them to determine around 230 of the top genes that would provide the most interesting variation in the dataset.

% the authors labeled each spot by its region in the brain which we decided to use as a target feature for a classifier. unfortunately the raw labels were unsuitable for training because there were 15 classes with high imbalance

Additionally, we discovered that the target feature consisted of 15 classes, yet most of the data was categorized under non-highly differentiable labels, and despite an abundance of classes, there was high class imbalance. The authors were using an ontology provided by the Allen Brain Atlas for adult mouse brains~\cite{Bioportal}. After referencing the original ontology, we discovered three high level classes that would optimally separate the data among the 15 classes: the brain stem, the cerebellum, and the cerebrum.

\subsection{Preprocessing}

\subsubsection{Validation set split}
As the data we were analysing was sampled based on their associative tissue slides, we couldn't simply use a randomized sampling approach to split the dataset into its respective modeling sets. We enacted a strategy to use a fraction of the total samples to be held out for validation. With 75 samples in total, we intended to use 7 samples as part of this holdout set. However, there was a clear class imbalance for the target classes. We plotted the samples against their total inherent class representation and annotated those which had representation above a specified threshold of at least 10 classes. Finally, we randomly selected 7 of the highly representative samples to be held out for validation.  %[Figure~\ref{fig:valset}]. 

% \begin{figure}[htp]
%     \centering
%     \includegraphics[width=9cm]{images/output.png}
%     \caption{Target class representation across all samples.}
%     \label{fig:valset}
% \end{figure}

\subsubsection{Filtration and Binning}
To round out the preprocessing, we filtered our feature set to only those genes which the authors for the source dataset identified. Furthermore, we categorized the 15 classes of the target feature into the three isolated classes we discovered through the ABA ontology~\cite{Bioportal}.

\subsubsection{Reformatting Spots to Graphs}
As the final step in our preprocessing, we used Pytorch Geometric to load the dataset as graphs rather than singular data points. We applied a Radius Graph transformation in order to capture and represent points in a grid like manner. The radius parameter was a tuned until we found an optimal arrangement of points within each graph.

\section{Results}
For our experimental models, we trained a Graph Convolution Network and three neural operator models~\ref{tab:Abbs}. All models were trained with a balanced class weighting. Both accuracy and F1 metrics were considered for evaluation, however the F1 metric was chosen as the primary evaluation metric as it is the preferred metric in evaluating highly imbalanced sets. Table~\ref{tab:ClassPerf} reflects the evaluation metrics for each model that we trained against this dataset. \\

\begin{table}[htp]
\setlength{\arrayrulewidth}{0.4mm}
\setlength{\tabcolsep}{9pt}
\renewcommand{\arraystretch}{1.4}
\fontsize{8}{11}\selectfont
\caption{Classification Accuracies and F1 scores for the Baseline and GNN models (n=10).}
\begin{tabular}{ 
|p{1.9cm}|p{1.7cm}|p{1.7cm}|p{0.8cm}|  }
\hline
\multicolumn{4}{|c|}{Classification Performance} \\
\hline
Baseline Models & Accuracy & F1-Score & \# of Params \\
\hline
LR & $48.96 \pm 0.00\%$    &$52.50 \pm 0.00\%$& {3} \\
FCN & $59.26 \pm 0.00\%$   &$58.20 \pm 0.00\%$& {5} \\
RF &$67.90 \pm 0.27\%$ &\cellcolor{yellow!25}$62.30 \pm 0.12\%$& {4} \\
XGB    &\cellcolor{yellow!25}$68.45 \pm 0.00\%$ &$61.79 \pm 0.00\%$& {0} \\
\hline
GNN Models & {} &{}& {} \\
\hline
SpatialKernel &$64.08 \pm 6.61\%$  &$54.16 \pm 2.40\%$& {2} \\
GraphSAGE    &$43.61 \pm 1.56\%$   &$44.09 \pm 2.64\%$& {3} \\
GCN & $52.18 \pm 3.35\%$   &$55.91 \pm 2.11\%$& {2} \\
GAT    &$50.31 \pm 3.09\%$ &$53.51 \pm 2.25\%$& {4} \\
GIN    &$53.99 \pm 2.27\%$ &$54.00 \pm 1.74\%$& {3} \\
SpatialGCN    &$62.61 \pm 5.72\%$  & $51.99 \pm 2.54$\%& {2} \\
GraphPDE   & \cellcolor{green!25}$67.63 \pm 1.35$\%    &\cellcolor{green} $71.06 \pm 0.59$\%& {5} \\
\hline
\end{tabular}
\label{tab:ClassPerf}
\end{table}

\subsection{Graph Convolutional Network}
\begin{figure}[htp]
    \centering
    \includegraphics[width=9cm]{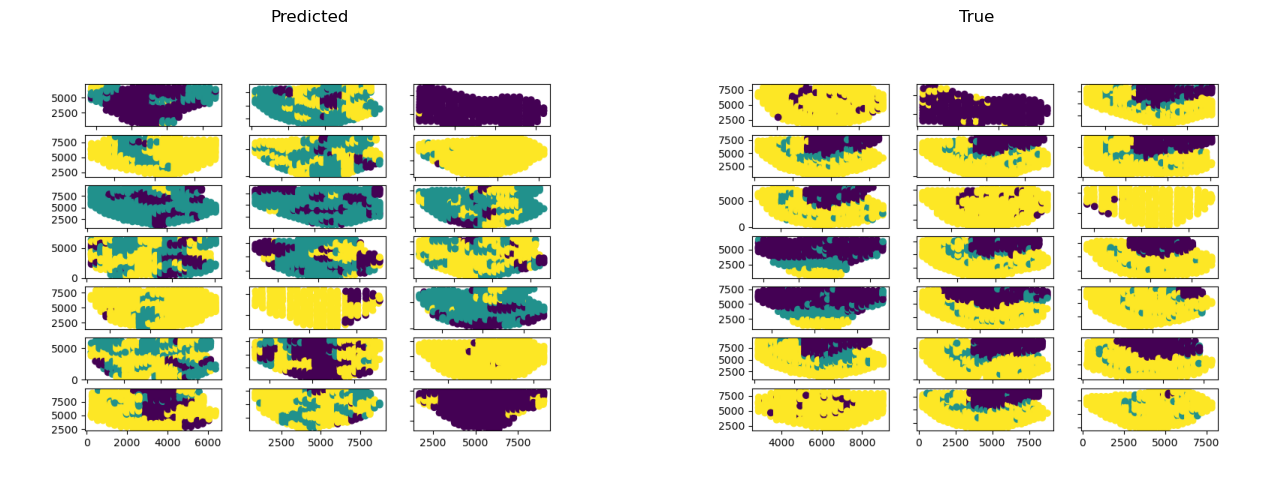}
    \caption{Predicted vs True class representation of the GCN model.}
    \label{fig:gcn}
\end{figure}

The GCN [Figure~\ref{fig:gcn}] has a very low performance when factoring in balanced class weights. It predicts all 3 classes for the most part.

\subsection{Neural Operators}

For the operator networks, we chose to test both simple and convoluted variations to validate the robustness of each. For our simple approach, we built a kernel network termed SpatialKernel which incorporated a Gaussian norm and linear layers. We further tested the simple approach by replacing the linear layers with GCN convolution layers for the SpatialGCN model. Finally, for the convoluted approach, we implemented the GraphPDE approach ~\cite{GraphPDE} with a graph network as the kernel layer.

\subsubsection{SpatialKernel}
\begin{figure}[htp]
    \centering
    \includegraphics[width=9cm]{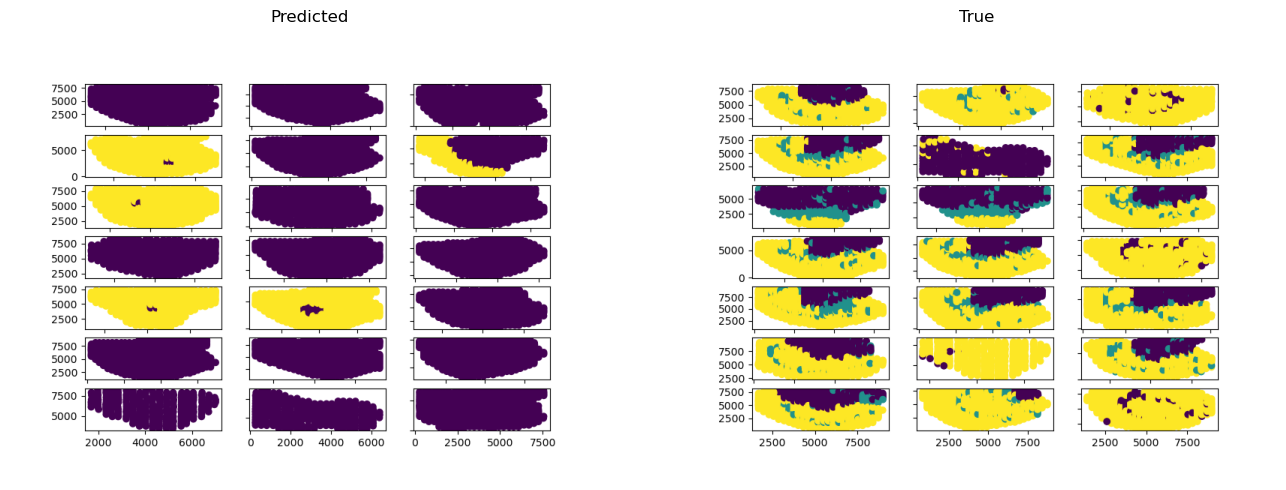}
    \caption{Predicted vs True class representation of the SpatialKernel model.}
    \label{fig:spat}
\end{figure}

The SpatialKernel model [Figure~\ref{fig:spat}] uses three linear layers, wherein a Gaussian kernel is computed on the positional features and multiplied to the X tensor before each linear layer. This model performs worse than the GCN, however it seems to be doing a better job in generalizing across the classes.

\subsubsection{SpatialGCN}
\begin{figure}[htp]
    \centering
    \includegraphics[width=9cm]{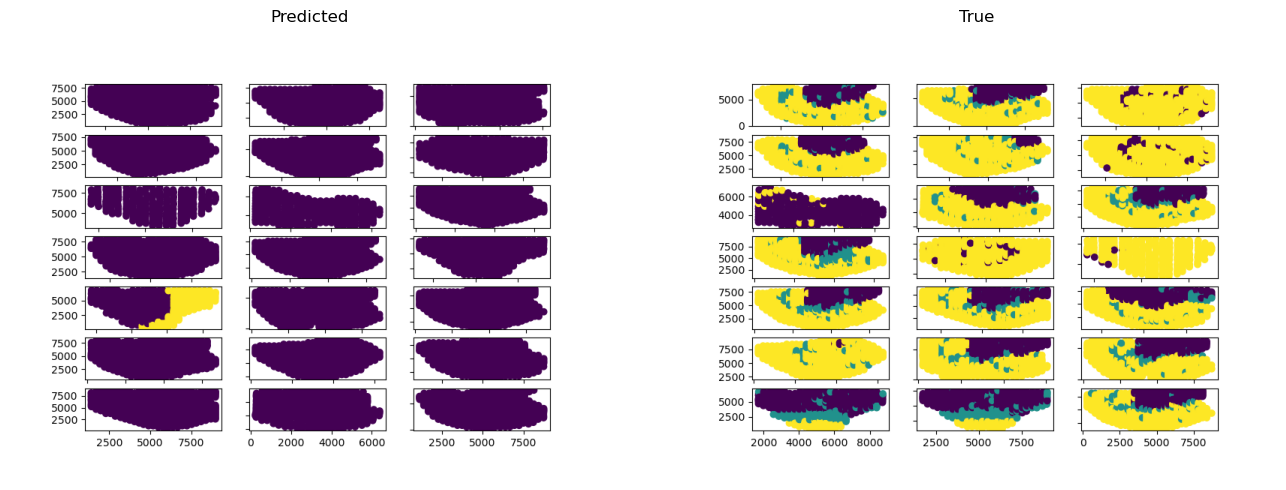}
    \caption{Predicted vs True class representation of the SpatialGCN model.}
    \label{fig:spatgcn}
\end{figure}

The SpatialGCN model [Figure~\ref{fig:spatgcn}] combines both the SpatialKernel and GCN such that each layer is a GCNConv layer, and a gaussian kernel is computed before each convolutional layer. This model shows better performance than the previous two models, however it is not predicting for the third class at all.

\subsubsection{GraphPDE}
\begin{figure}[htp]
    \centering
    \includegraphics[width=9cm]{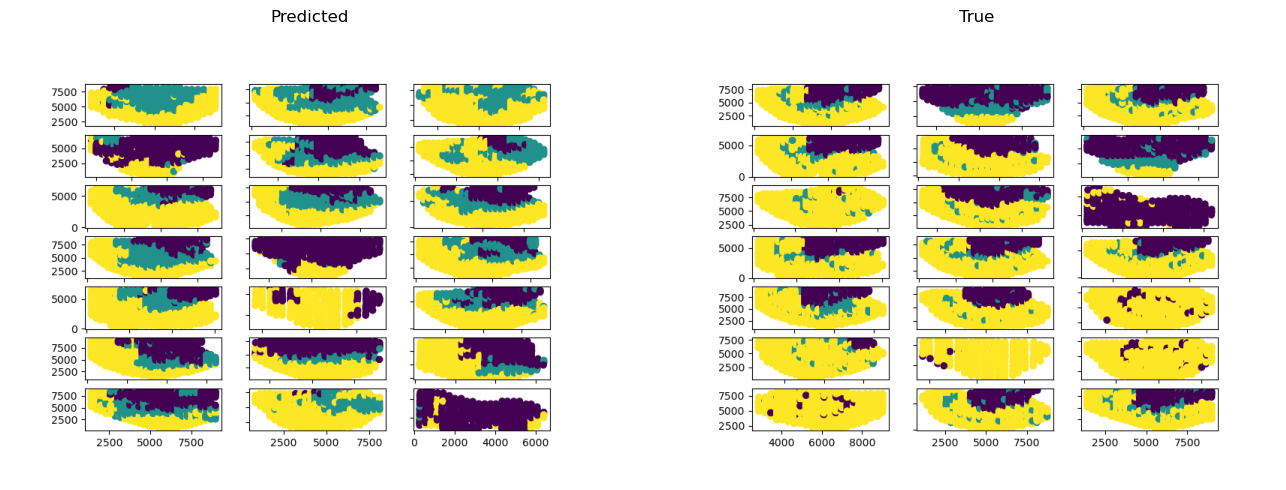}
    \caption{Predicted vs True class representation of the GraphPDE model.}
    \label{fig:gpde}
\end{figure}

The GraphPDE network [Figure~\ref{fig:gpde}] required some additional customization in the data loading procedures by incorporation of edge attributes. Following this procedural addition, we trained the foundation GraphPDE model against the dataset using only six hidden layers. This model takes in more parameters compared to the other experimental models. This model seems to be the most reliable both in terms of performance and generalization across classes. It has the highest reported accuracy and F1-score than the other graph network methods. It is slightly better than the XGBoost model (the baseline model to beat) in accuracy and far better than it or any other in F1-score performance. 

\subsection{Other Methods}
Other models for which we considered training against the dataset were supplementary. They were merely used to demonstrate the robustness of the GraphPDE model against high caliber graph neural network models. Discussion on their performance can be found in the Supplementary Information section.

\section{Conclusion}
In this paper, we demonstrated the feasibility of using graph learning techniques to classify mouse brain regions with a particular highlight into the appication of neural operators. The GraphPDE model proved to be robust both in performance and generaralization across classes. It outperformed all the baselines and similar caliber graph networks with an F1-score of 72\%. This research presents graph neural operators as valid performers in supervised classification tasks on spatial transcriptomics data.

\section{Discussion and Future Work}
The study could be improved in multiple facets. First, it was a fast tracked research endeavor considering a limited timeline. Many of the GNN models of interest employed some semblance of early fusion tactics incorporating multimodal data. With additional time permitting, we would consider additional data fusion techniques to compare against neural operators.

Second, the dataset at the point of discovery was one of few nearly perfect curated sets. This dataset has sufficient samples, spots, and genes, however it lacks in generalization. The data comes from one mouse brain. Replicating results gathered to other mouse brains, should they be gathered, may affect the final performance of each model. In a more general sense, we may consider applying this technique to datasets that are not of mouse brains, such as the mouse spinal cord dataset~\cite{MouseSpine}, and even cross-species analysis in human anatomy datasets.

Third, the implemented GraphPDE model was shallow with a depth of only six layers. With the benefit of additional time, we would investigate the space of optimal hyperparameters across all models, and further determine whether a deeper GraphPDE network could provide stronger performance. Furthermore, on the note of hyperparameters, there were many models for which we didn't parameterize certain values. For example, we could have considered parameterizing the sigma constant in the SpatialKernel model to allow learnability for the most optimal value.

Another method we may consider in the future is late fusion techniques. The methods we utilized reflected early fusion techniques where the data is treated in a multimodal fashion through a single model pass. For future works, we may try to train separate classifiers: one for the feature information and the other either for the positional embeddings or an image classifier for microscope images of the respective samples. Next, we would consider feeding outputs from both of these into a single aggregator model to derive a final output.  

Finally, we employed one technique out of many possible neural operator approaches in this research. Other neural operator techniques incorporate state of the art architectures, for example, transformers as a kernel approximation as opposed to the graph network used in the GraphPDE. We may consider using this and other high caliber neural operator approaches for further research.

{\footnotesize \bibliographystyle{acm}
\bibliography{st}}

\section{Appendix}

\subsection{Abbreviations}
\label{tab:Abbs}
\begin{center}
\begin{tabular}{ c c }
 ABA & Allen Brain Atlas \\
 FCN & Fully Connected Network \\
 GAT & Graph Attention Network \\  
 GIN & Graph Isomorphism Network \\
 GraphPDE & Graph Neural Operator \\
 GNN & Graph Neural Network \\
 LR & Logistic Regressor \\
 RF & Random Forest \\ 
 XGB & XGBoost model
\end{tabular}
\end{center}

\subsection{Supplementary Information}
\subsubsection{Graph Attention Network}
\begin{figure}[htp]
    \centering
    \includegraphics[width=9cm]{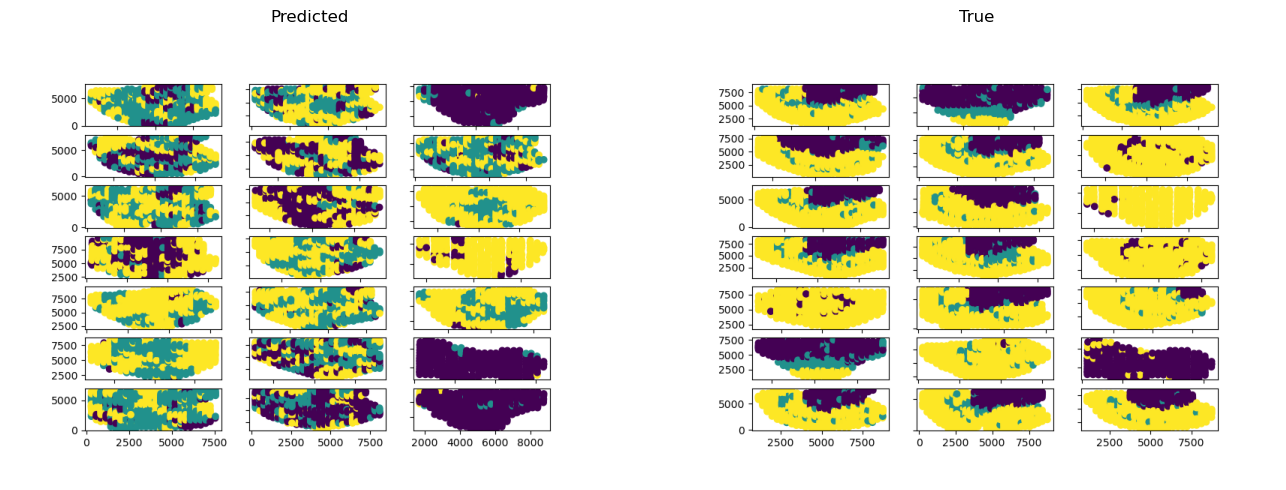}
    \caption{Predicted vs True class representation of the GAT model.}
    \label{fig:gat}
\end{figure}

The GAT [Figure~\ref{fig:gat}] is not comparable to the top performers in terms of accuracy, nor is it predictive towards all three classes. Furthermore it takes longer to train, so it is not a worthy tradeoff.

\subsubsection{GraphSAGE}
\begin{figure}[htp]
    \centering
    \includegraphics[width=9cm]{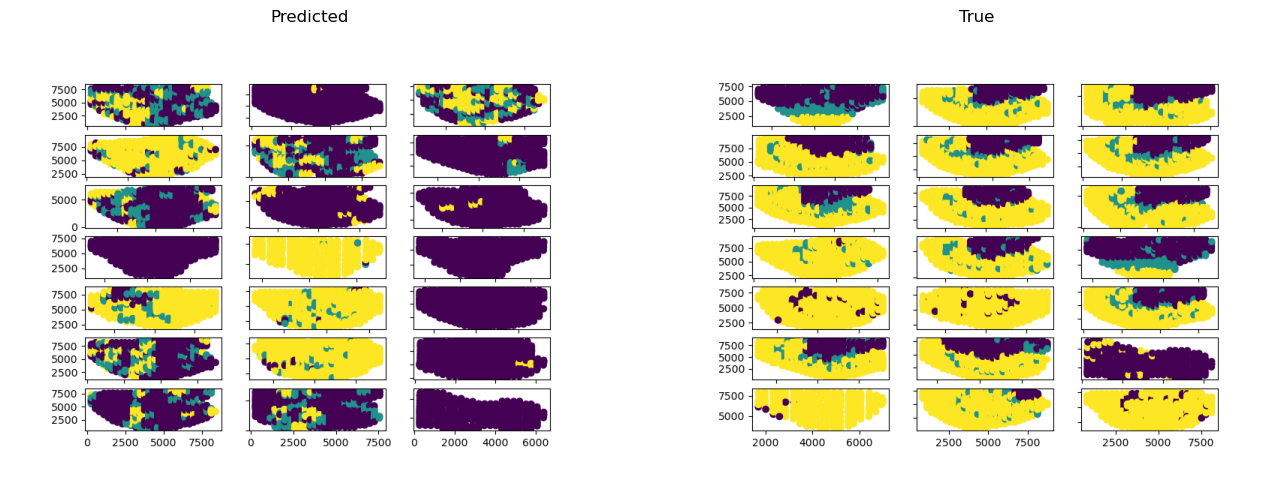}
    \caption{Predicted vs True class representation of the GraphSAGE model.}
    \label{fig:gsage}
\end{figure}

The GraphSAGE [Figure~\ref{fig:gsage}] network performs worse than the GAT model and slightly better than the simple SpatialKernel model. It is not highly predictive towards all classes.

\subsubsection{Graph Isomorphism Network}
\begin{figure}[htp]
    \centering
    \includegraphics[width=9cm]{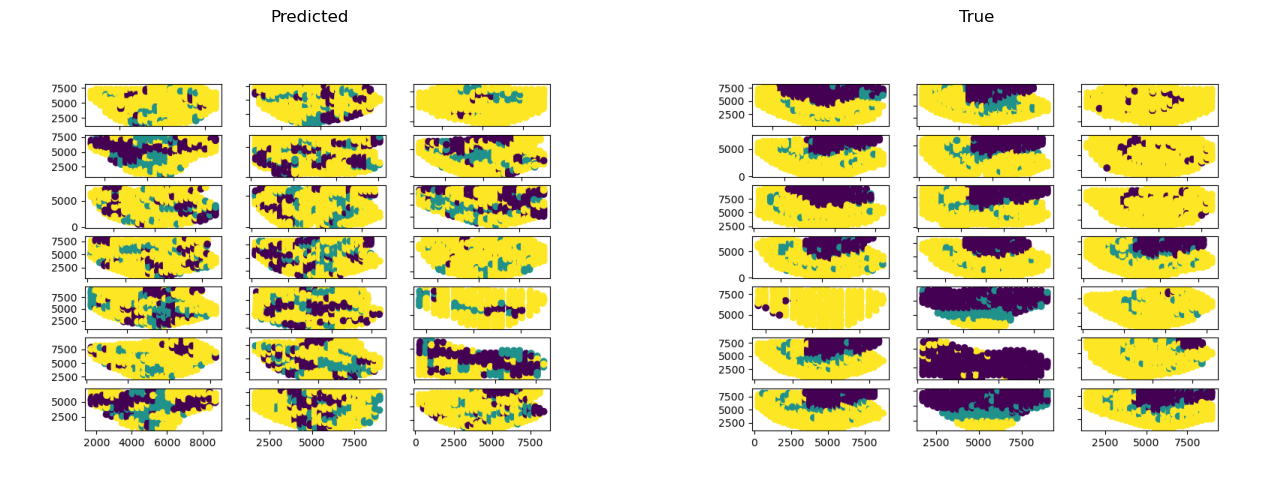}
    \caption{Predicted vs True class representation of the GIN model.}
    \label{fig:gin}
\end{figure}

This network [Figure~\ref{fig:gin}] performs better than the GAT and the GraphSAGE models, which isn't very telling among the breadth of GNN model performers, yet is comparable to the suite of actual graph baselines. It is more accurate than the GCN and the loss is minimal compared to other models. It lacks in terms of representation of all three classes, and both with low accuracy and F1-score, it is not a worthy comparator against the GraphPDE.

\end{document}